# PsycholexTherapy: Simulating Reasoning in Psychotherapy with Small Language Models in Persian

Mohammad Amin Abbasi[1], Hassan Naderi[2*]

Department of Computer Engineering Iran University of Science and Technology, Tehran, Iran

[1]`m_abbasi1378@comp.iust.ac.ir`
[2]`naderi@iust.ac.ir`

This study presents PsychoLexTherapy, a framework for simulating psychotherapeutic reasoning in Persian using small language models (SLMs). The framework tackles the challenge of developing culturally grounded, therapeutically coherent dialogue systems with structured memory for single-turn and multi-turn interactions in underrepresented languages. To ensure privacy and feasibility, PsychoLexTherapy is optimized for on-device deployment, enabling use without external servers. Development followed a three-stage process: (i) assessing SLMs' psychological knowledge with PsychoLexEval; (ii) designing and implementing the reasoning-oriented PsychoLexTherapy framework; and (iii) constructing two evaluation datasets—PsychoLexQuery (real Persian user questions) and PsychoLexDialogue (hybrid simulated sessions)—to benchmark against multiple baselines. Experiments compared simple prompting, multi-agent debate, and structured therapeutic reasoning paths. Results showed that deliberate model selection balanced accuracy, efficiency, and privacy. On PsychoLexQuery, PsychoLexTherapy outperformed all baselines in automatic LLM-as-a-judge evaluation and was ranked highest by human evaluators in a single-turn preference study. In multi-turn tests with PsychoLexDialogue, the long-term memory module proved essential: while naïve history concatenation caused incoherence and information loss, the full framework achieved the highest ratings in empathy, coherence, cultural fit, and personalization. Overall, PsychoLexTherapy establishes a practical, privacy-preserving, and culturally aligned foundation for Persian psychotherapy simulation, contributing novel datasets, a reproducible evaluation pipeline, and empirical insights into structured memory for therapeutic reasoning.

## 1 INTRODUCTION

Large language models (LLMs) have recently emerged as powerful tools for natural language understanding and generation, and their use has expanded into sensitive domains such as mental health[1, 2]. In this context, LLMs are increasingly applied to support mental-health–oriented dialogue, offering opportunities for empathetic communication, mental health education, and therapeutic support. Early work on empathetic dialogue systems demonstrated that exposure to emotionally grounded corpora improves the perceived empathy and engagement of conversational agents. In parallel, the development of benchmarks and datasets for empathetic response generation has enabled more systematic evaluation and comparison of models[3].

Clinical and digital health research has also investigated therapy-oriented chatbots. Early randomized trials of systems aligned with cognitive behavioral therapy (CBT) reported short-term improvements in depressive and anxiety symptoms, indicating the practical potential of automated dialogue systems for mental health[4]. However, these findings are accompanied by cautionary notes on safety, user trust, cultural appropriateness, and the sustainability of such interventions. Reviews in digital psychiatry and global mental health emphasize that cultural and contextual adaptation is not an optional feature but a requirement for effectiveness and acceptability, particularly in underrepresented languages and communities[5].

---

[*] Corresponding author

At the technical level, research on memory-augmented LLMs has introduced explicit mechanisms for long-horizon coherence, personalization, and context retention[6, 7]. Such advances are crucial for therapy-like interactions that require consistent reasoning across multiple turns. In parallel, trends in on-device deployment of smaller models are emerging as promising strategies to protect privacy, reduce latency, and enable feasibility in sensitive domains like psychotherapy[8, 9]. Together, these streams highlight both the promise and the challenges of applying LLMs in mental health contexts.

Despite substantial progress, several gaps remain in the current literature. First, there is limited understanding of the psychological knowledge encoded in small language models, raising questions about whether such models can support reasoning-oriented therapeutic dialogue. Second, most existing dialogue frameworks lack a dedicated design for simulating psychotherapeutic reasoning, relying instead on generic prompts or surface-level empathy rather than structured, workflow-driven approaches that resemble professional therapeutic practice. Third, evaluation efforts have been largely confined to isolated or short-turn conversations, without systematically assessing performance in real-world single-turn queries or sustained multi-turn therapy-like dialogues. These limitations restrict our ability to advance beyond ad-hoc empathetic response generation toward dialogue systems that are coherent, culturally grounded, and reasoning-capable in psychotherapeutic contexts.

This paper introduces PsychoLexTherapy, a framework for simulating psychotherapeutic reasoning in Persian with small, locally deployable language models. The central research question is: Can structured therapeutic reasoning paths and explicit long-term memory mechanisms, implemented on compact models, yield coherent, empathetic, and culturally aligned dialogue that surpasses baseline prompting methods?

To address this, the study pursues three objectives:
- Psychological knowledge assessment: Evaluate whether small models possess the foundational psychological knowledge required for reasoning in therapy-like settings.
- Framework design: Develop a structured framework that simulates psychotherapeutic reasoning paths with small language models.
- Empirical evaluation: The framework is evaluated against competitive baselines in both single-turn queries and multi-turn therapeutic dialogues, focusing on empathy, coherence, cultural alignment, and personalization, using both automatic metrics and expert human assessment by psychologists.

Correspondingly, the study makes three key contributions. First, it introduces PsychoLexEval, a dataset for systematically probing the psychological knowledge of small models, ensuring that the chosen base model meets a minimum threshold of competence. Second, it presents PsychoLexTherapy, a reasoning-oriented therapeutic framework designed for local deployment, incorporating structured reasoning paths and a long-term memory module. Third, it contributes two evaluation resources, PsychoLexQuery, a collection of real Persian user questions, and PsychoLexDialogue, a dataset of simulated therapeutic conversations. These resources provide the foundation for comprehensive evaluation and demonstrate the empirical benefits of the proposed framework over strong prompting-based baselines.

## 2 RELATED WORK

### 2.1 Empathetic Dialogue and Support-Oriented Datasets

Empathetic dialogue has been a foundational area for studying affective capabilities in conversational agents[10]. The EmpatheticDialogues dataset, containing 25k conversations grounded in emotional situations, provided one of the earliest large-scale benchmarks for evaluating empathy in open-domain dialogue. Subsequent research demonstrated that models fine-tuned on this corpus could generate more emotionally attuned responses and increase user engagement compared to generic pre-trained models[11]. Building on this foundation, ESConv introduced the Emotional Support Conversation task, which explicitly mapped dialogues to Helping Skills Theory and annotated strategy use, thus moving from surface-level empathy to structured support strategies[12]. This shift highlighted that empathy alone is insufficient for therapeutic quality—conversation agents also require reasoning strategies aligned with counseling practice[4].

In parallel, researchers explored mental-health–specific corpora. PsyQA, a large-scale Chinese counseling Q&A dataset, systematically annotated support strategies to probe model reasoning in therapy-like interactions[13]. Such culturally specific corpora illustrate the importance of grounding data in local contexts rather than relying exclusively on English-language resources. To address data scarcity, synthesis approaches have emerged. For example, SMILECHAT generated over 55k multi-turn counseling dialogues via scripted pipelines and LLM generation, enabling controlled experimentation at scale[13]. Complementary resources such as CounselChat, where licensed therapists provide written answers to real user-submitted questions, emphasize ecological validity by grounding evaluations in authentic counseling queries[14]. Together, these efforts mark a progression from general empathy benchmarks to domain-specific, strategy-annotated, and culturally grounded datasets—a trajectory that directly motivates the design of Persian resources in our study.

### 2.2 Therapy-Oriented Chatbots and Clinical Evidence

Alongside data resources, clinical and digital-health research has investigated the real-world impact of therapy-oriented chatbots. Early randomized controlled trials of systems such as Woebot, designed around principles of cognitive behavioral therapy, reported short-term reductions in symptoms of depression and anxiety in college students. These findings demonstrated the feasibility of automated dialogue systems to deliver therapeutic benefit[15]. However, they also underscored limitations: effects were short-term, long-term safety and trust remained untested, and users expressed concerns about the authenticity of chatbot empathy. Later studies with language-specific or culturally adapted CBT bots (e.g., a Polish-language system for depressive symptoms) replicated partial benefits, reinforcing the importance of cultural alignment.

Systematic reviews and meta-analyses provide further perspective. They emphasize that while therapy-oriented chatbots can increase accessibility, risks include inaccurate advice, dependency, or misaligned expectations. Ethical frameworks caution that unsupervised deployment may cause harm unless accompanied by professional oversight, transparent limitations, and strict privacy protections[2]. Thus, the literature reveals both promise and peril: while chatbots can be clinically useful, their design must go beyond superficial empathy and integrate therapeutic reasoning, safety protocols, and cultural fit.

### 2.3 Cultural and Contextual Adaptation in Digital Mental Health

Cultural adaptation has emerged as a defining factor for digital mental-health interventions. Reviews in digital psychiatry argue that effectiveness depends on tailoring to local languages, idioms of distress, and cultural models of therapy[16]. Taxonomies of adaptation components—ranging from linguistic translation to contextualizing therapy strategies—show that interventions adapted for underrepresented groups achieve greater acceptance, adherence, and effectiveness compared to generic versions[17]. In practice, adaptation involves not just language but also aligning therapeutic strategies with local social norms and expectations of help-seeking.

Recent implementation-science perspectives stress the need for structured frameworks to guide adaptation in digital tools. This includes identifying when adaptation is necessary (e.g., during content design, interface design, or evaluation) and how to systematically incorporate feedback from both target users and clinicians. In underrepresented languages such as Persian, the absence of tailored digital-mental-health frameworks represents a critical gap. Without such adaptation, systems risk irrelevance or rejection, regardless of technical sophistication. This directly motivates our focus on Persian psychotherapy simulation and culturally grounded evaluation.

### 2.4 Memory-Augmented Conversational Agents

Therapy-like dialogue requires consistency across sessions, recalling prior disclosures, and maintaining coherent therapeutic reasoning over time. Standard LLMs, constrained by context window size, often fail in long-horizon conversations. Research in memory-augmented conversational agents seeks to overcome this limitation. Systems such as MemGPT[18] introduce externalized memory modules that allow agents to dynamically store and retrieve past information. Approaches like Memlong[7] decouple short-term working memory from long-term episodic memory, enabling scalable reasoning across sessions.

Benchmarks such as LoCoMo[6] (Long-Context Modeling) and LongMemEval[19] explicitly test agents' ability to remember and reason across extended dialogues. Surveys in this area consolidate design patterns into implicit memory (hidden in model parameters), explicit memory (structured external stores), and agentic memory (meta-level reasoning about memory)[20].

The consensus is clear: structured memory mechanisms are essential for dialogue domains requiring coherence and personalization. For psychotherapy, this is particularly critical, as lapses in recall or inconsistency can undermine trust and therapeutic alliance. Our framework extends this line by integrating structured memory with culturally grounded therapeutic reasoning paths, moving beyond generic memory augmentation to therapy-specific reasoning.

### 2.5 Small and On-Device Models for Privacy

While large cloud-based models dominate academic benchmarks, sensitive applications like psychotherapy require privacy-preserving alternatives. The emergence of small language models offers a promising path. A recent systematic evaluation of over 60 SLMs (e.g., Gemma[21], Qwen[22]) demonstrated their competitiveness on reasoning and language tasks, while being lightweight enough for deployment on personal devices[23, 24]. Industry platforms are similarly expanding on-device inference support, enabling applications that never leave the user's local environment[25]. This shift to edge deployment is crucial in mental health, where confidentiality and user control are paramount. By designing PsychoLexTherapy specifically for SLMs, we align with this trend, ensuring feasibility in low-resource settings and eliminating reliance on external servers. Unlike prior work that assumes access to large, cloud-hosted models, our study explicitly demonstrates that reasoning-aligned psychotherapy simulation can be achieved with compact, locally deployable models.

## 2.6 Safety, Ethics, and Professional Oversight

The ethical stakes of therapy-oriented dialogue are high. Reviews consistently emphasize risks such as misinformation, inappropriate responses to crisis disclosures, and potential psychological harm[26]. Professional guidelines recommend embedding clear guardrails, transparent disclaimers, and escalation mechanisms into digital mental-health tools[27]. Empirical evaluations of chatbots reveal variability in performance: while some systems provide useful psychoeducational content, others fail basic safety checks. This underscores the necessity of domain-specific evaluation frameworks that go beyond generic dialogue metrics to capture empathy, coherence, therapeutic alignment, and cultural fit[28].

Our work follows these recommendations by defining explicit evaluation dimensions grounded in psychological and linguistic criteria, validated with human annotators. By integrating structured evaluation with novel datasets, we address calls for rigorous, reproducible frameworks in digital mental health.

## 2.7 Positioning of This Work

In summary, four gaps remain in the literature: (i) limited exploration of reasoning-oriented therapeutic dialogue in underrepresented languages; (ii) lack of Persian psychotherapy datasets; (iii) limited use of structured long-term memory; and (iv) few privacy-preserving, on-device agents built on SLMs. To address these, we introduce PsychoLexEval, PsychoLexQuery, and PsychoLexDialogue as the first Persian psychotherapy datasets, together with the PsychoLexTherapy framework. This pipeline establishes a culturally aligned, memory-enabled approach and extends the reach of digital mental-health research to Persian.

## 3 METHODOLOGY

The research followed a stepwise methodology designed to progressively build and evaluate a psychotherapy-oriented dialogue framework in Persian. The design consisted of three interconnected stages, each addressing a distinct layer of the problem.

The first stage examined whether small language models possess sufficient domain knowledge to support reasoning in psychotherapy contexts. For this purpose, we developed the PsychoLexEval dataset, consisting of multiple-choice psychology questions covering core therapeutic concepts. By benchmarking candidate models on this resource, we ensured that the eventual base model met a minimum threshold of psychological competence before being deployed in therapy-like simulations.

The second stage focused on the development of the PsychoLexTherapy framework, the central contribution of this study. The framework integrates structured therapeutic reasoning path with dialogue management and introduces a long-term memory module (MemoBase[†]) for coherence and personalization across sessions. This stage formalized the reasoning process of therapeutic dialogue, moving beyond generic prompting toward a design that explicitly models psychotherapeutic reasoning.

In the final stage, the framework was evaluated against competitive baselines in both single-turn and multi-turn settings. For single-turn evaluation, we introduced PsychoLexQuery, a dataset of real Persian user questions drawn from mental-health forums, to measure empathy, coherence, and cultural alignment. For multi-turn evaluation, we created PsychoLexDialogue, a hybrid dataset combining scripted and LLM-assisted dialogues to approximate realistic therapy sessions. Evaluations included both quantitative metrics and human judgment, allowing for a robust assessment of the framework's effectiveness relative to strong baselines.

---

[†] www.memobase.io

By organizing the research in this sequential manner—knowledge assessment, framework design, and empirical evaluation—we ensured that the final system rested on a validated knowledge foundation, incorporated therapy-specific reasoning mechanisms, and underwent rigorous empirical testing in both controlled and ecologically valid scenarios.

### 3.1 PsychoLexEval

The first stage of this study focused on assessing whether small language models possess sufficient psychological knowledge to serve as a foundation for reasoning-oriented therapeutic dialogue. This step was necessary because language models, trained primarily on large-scale general corpora, often lack the domain-specific accuracy and depth required for sensitive applications such as psychotherapy. Without such validation, reliance on their outputs could result in content that is factually incorrect or clinically inappropriate.

To address this gap, we designed the PsychoLexEval dataset, a domain-specific resource in Persian for probing psychological knowledge. The dataset was structured as multiple-choice questions, enabling systematic evaluation of models' ability to recall and reason about key psychological concepts. Source materials were drawn from a diverse set of reliable and domain-relevant references, including graduate entrance examinations in psychology, professional recruitment tests, verified online resources, and additional items generated with GPT-4o inspired by authoritative Persian psychology textbooks.

To ensure data quality, a multi-step curation and validation process was applied. This included: (i) reviewing all items for accuracy and relevance, (ii) removing incomplete, ambiguous, or low-quality questions, (iii) verifying that each question contained four valid answer choices, and (iv) eliminating content with potential copyright concerns. After this filtering process, the dataset was finalized with 3,430 multiple-choice questions, covering a broad spectrum of psychological subfields such as clinical, cognitive, developmental, and social psychology.

The primary purpose of PsychoLexEval was twofold: (1) to benchmark candidate small language models for their baseline psychological competence, thereby guiding base model selection for subsequent framework development; and (2) to provide a reusable, culturally grounded resource for evaluating Persian-language models in specialized domains.

### 3.2 PsychoLexQuery

To evaluate the proposed framework in single-turn settings, we constructed the PsychoLexQuery dataset. The primary purpose of this resource was to assess whether responses generated by the framework reflect not only textual reproduction but also empathetic understanding, emotional reflection, and therapy-oriented reasoning when confronted with real-world user questions.

Data were collected through web crawling and manual extraction of user-generated content from Persian-language psychology forums and consultation platforms, including EhyaCenter[‡], Moshaverfa[§], and Simiaroom[**]. These platforms provide open spaces where individuals publicly share psychological concerns and licensed psychologists or counselors respond. Using such sources offered two advantages: (i) the authenticity of the data, as they directly capture the lived experiences and concerns of Persian-speaking users; and (ii) the cultural richness of the material, which encodes linguistic subtleties, family dynamics, and societal expectations specific to the Iranian context.

---

[‡] ehyacenter.com
[§] moshaverfa.com
[**] simiaroom.com

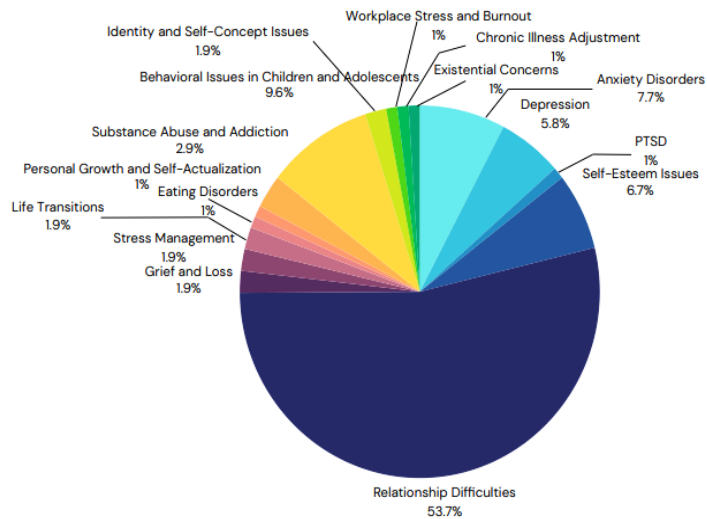

Figure 1: The distribution of topics in PsycholexQuery.

In total, approximately 4,000 user questions were collected, covering a broad spectrum of psychological themes such as anxiety, depression, family conflict, academic pressure, workplace stress, and relationship issues. These naturally occurring queries provided a valuable benchmark for testing cultural sensitivity and therapeutic alignment in generated responses.

Given the sensitivity of mental-health data, extensive anonymization and preprocessing procedures were applied to ensure ethical usability. All items were manually reviewed, with the following key steps implemented:
- Replacing real names with generic placeholders (e.g., "Person A", "Participant B").
- Generalizing geographic identifiers to broader terms such as "residential area" or "large city".
- Abstracting temporal references to month or year, depending on context.
- Removing unique identifiers including phone numbers, emails, and social media handles.
- Generalizing context-specific attributes such as job titles or project names (e.g., "CEO of Company X" → "senior manager").

Through these steps, PsychoLexQuery retained its cultural authenticity while ensuring ethical standards of confidentiality and participant privacy. The dataset thus provides a secure and context-sensitive foundation for evaluating single-turn therapeutic reasoning in Persian. The distribution of topics is shown in Figure 1.

### 3.3 PsycholexDialoge

The PsychoLexDialogue dataset was developed to provide a robust testbed for evaluating multi-turn psychotherapy simulations, with particular emphasis on examining the contribution of structured long-term memory in dialogue coherence and personalization. Unlike single-turn resources, which capture isolated user questions, multi-turn dialogues are essential to model the evolving dynamics of therapy, where each turn builds on prior disclosures and therapeutic reasoning.

Longitudinal therapeutic interactions require continuity: therapists recall prior concerns, track recurring emotional themes, and adapt interventions accordingly. Off-the-shelf datasets rarely meet these requirements, especially in underrepresented languages such as Persian. To fill this gap, PsychoLexDialogue was designed with three guiding principles:

- Psychological Depth – dialogues should represent multi-layered therapeutic processes rather than single-problem interactions.

---

**Client Question:**
"Hello,
I struggle with assertiveness in tasks that require interaction and collaboration with others. Sometimes, being too available, compromising too much, accepting others' mistakes, and constantly giving in make me feel exhausted and worthless. In response, I become stubborn as a way to make those around me realize how they've taken advantage of my flexibility.
I want to understand the difference between stubbornness and assertiveness. Also, I need guidance on how to recognize when I should be assertive. I naturally tend to be soft and flexible, but I feel that it's not always the right approach."

**Client Profile:**
```
{
  "emotional_themes": ["frustration", "insecurity", "exhaustion", "confusion", "desire for assertiveness"],
  "key_psychological_issues": ["lack of assertiveness", "fear of being taken advantage of", "self-worth issues", "difficulty in setting boundaries"],
  "past_experiences": ["experiences of being overly accommodating in relationships", "feelings of being undervalued or unappreciated"],
  "patterns_and_behaviors": ["over-accommodating behavior", "difficulty in asserting needs", "oscillation between flexibility and stubbornness"],
  "desired_outcome": "guidance on distinguishing between assertiveness and stubbornness, and strategies for being more assertive",
  "contextual_factors": ["the user is likely in a collaborative work environment", "the user may have a tendency to prioritize others' needs over their own"]
}
```

---

Figure 2: Example of a client question and its generated profile.

Table 1: Frequency of the main emotional themes extracted from the PsychoLexQuery dataset

| Emotional Theme | Frequency |
|---|---|
| Frustration | 2,914 |
| Sadness | 2,124 |
| Anxiety | 1,306 |
| Fear | 1,228 |
| Insecurity | 1,224 |
| Confusion | 1,150 |
| Worry | 678 |
| Helplessness | 415 |
| Hopelessness | 266 |
| Guilt | 216 |

- Cultural Authenticity – conversations should reflect Persian linguistic norms, family structures, and social contexts.
- Evaluative Utility – the dataset should explicitly enable testing of memory-enabled versus memory-free systems.

The foundation of PsychoLexDialogue was built from PsychoLexQuery. Each query was expanded into a structured user profile consisting of:

- Emotional themes (e.g., frustration, sadness, fear, insecurity).
- Core psychological issues (e.g., trust difficulties, abandonment fears, self-worth struggles).
- Past experiences (e.g., family conflict, trauma, unmet needs).
- Cognitive/behavioral patterns (e.g., rumination, avoidance, reassurance-seeking).
- Therapeutic goals (e.g., coping strategies, emotional acceptance, confidence-building).
- Contextual factors (e.g., workplace stress, societal expectations).

These profiles provided the psychological scaffolding for dialogue generation, ensuring that conversations unfolded as therapeutic trajectories rather than unrelated exchanges. An example of a generated client profile is in Figure 2. Also, the distribution of the main emotional themes identified in the PsychoLexQuery dataset is presented in Table 1.

An analysis of emotional frequencies confirmed the dominance of negative affective states, aligning with real-world concerns in Persian-speaking populations. For instance, frustration appeared in 2,914 instances, sadness in 2,124, and anxiety in 1,306, followed by fear, insecurity, confusion, and hopelessness. The co-occurrence of multiple emotional themes (average 3–5 per case) reflected the complex, layered nature of psychological distress.

Each profile was expanded into a five-stage therapeutic narrative aligned with person-centered therapy (PCT):
1. Establishing trust and rapport.
2. Empathic listening and emotional reflection.
3. Encouraging free exploration of thoughts and feelings.
4. Supporting personal growth and cognitive restructuring.
5. Summarizing insights and planning future steps.

This stage ensured logical progression and emotional continuity across sessions, replicating how therapy unfolds in practice.

Narratives were first operationalized as scripted dialogues, where therapist and client turns were explicitly designed to follow the therapy flow. Scripted dialogues guaranteed thematic consistency, inclusion of therapeutic strategies, and coverage of core issues. However, they also risked mechanical delivery, with therapist responses sometimes sounding overly prescriptive and client turns lacking variability.

To address these shortcomings, we introduced a hybrid simulation layer where two LLM-based agents dynamically enacted the therapist and client roles:
- The therapist agent was guided by structured prompts enforcing PCT principles (empathy, open-ended questioning, reflection).
- The client agent drew from the predefined profile, incorporating natural variation such as ambiguity, defensiveness, contradiction, or sudden emotional shifts.

Each turn began with a scripted backbone, which was then refined through agent interaction. This hybrid cycle introduced realistic variability while retaining narrative coherence. Unlike free agent–agent simulations, which often drift into incoherence, our design anchored dialogues in a scripted trajectory but enhanced them with nuanced, context-aware adaptation.

The dialogue generation process followed an iterative four-step cycle:
- Script initiation: Dialogue turn begins from the predefined narrative.
- Therapist adaptation: Therapist agent revises the scripted response, incorporating client's latest state for emotional realism.
- Client response: Client agent generates context-sensitive reactions, including hesitation, contradictions, or escalated affect.
- Cycle continuation: Interaction proceeds to the next scripted step, balancing stability with flexibility.

This method yielded dialogues that were coherent, context-sensitive, and emotionally authentic.

The final version of PsychoLexDialogue included:
- 3,400 dialogues across 16 thematic categories.
- Average of 10–14 turns per session, sufficient to evaluate memory over extended interactions.
- Most common categories: relationship issues, anxiety, and self-esteem problems.

- Frequent emotional themes: hopelessness, sadness, anxiety, fear, insecurity, confusion.
- Average of 3–5 co-occurring emotions per dialogue, reflecting psychological complexity.

PsychoLexDialogue was specifically designed to test the effectiveness of the MemoBase long-term memory module within PsychoLexTherapy. By comparing system performance in memory-enabled vs. memory-free settings, it allows for controlled evaluation of memory's role in empathy, coherence, personalization, and cultural fit. More broadly, it provides the first Persian multi-turn psychotherapy dataset, offering a valuable benchmark for future research in underrepresented languages. Figure 3 illustrates the development pipeline of the PsychoLexDialogue dataset.

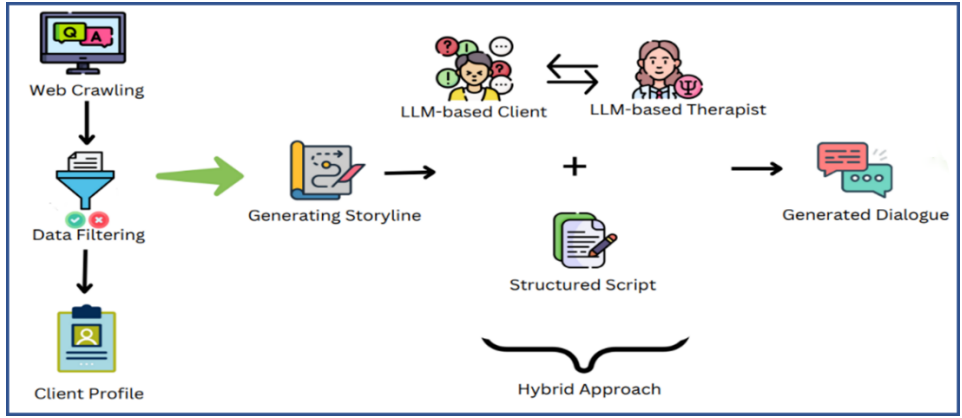

Figure 3: Development pipeline of the PsychoLexDialogue dataset.

### 3.4 PsychoLexTherapy Framework

Building on the results of the initial model selection stage, we designed and implemented the PsychoLexTherapy framework. The primary goal of this framework is to simulate the reasoning processes of human therapists across multiple psychotherapeutic approaches, including Cognitive Behavioral Therapy, Reality Therapy, and Person-Centered Therapy.

In real therapeutic practice, a psychologist typically follows a structured reasoning process when responding to a client: (i) identifying the core components of the client's message (thoughts, emotions, behaviors), (ii) analyzing these components through the lens of a specific therapeutic model, and (iii) providing an appropriate response or intervention. PsychoLexTherapy operationalizes this reasoning process by encoding each therapeutic approach as a stepwise reasoning path.

- In CBT reasoning paths, the system focuses on detecting dysfunctional thoughts and mapping their emotional and behavioral consequences.
- In RT reasoning paths, it evaluates the client's choices and their alignment with fundamental psychological needs, highlighting areas of responsibility and agency.
- In PCT reasoning paths, the emphasis is on empathy, emotional reflection, and creating a psychologically safe space for free self-expression.

By structuring these processes into explicit reasoning paths, responses generated by the system are not limited to surface-level mimicry of empathetic language. Instead, they are grounded in the internal logic of the therapeutic framework, producing outputs that are more structured, clinically meaningful, and evaluable.

This design ensures that PsychoLexTherapy not only generates fluent and natural responses but also provides therapeutically coherent interactions that can be assessed according to professional standards.

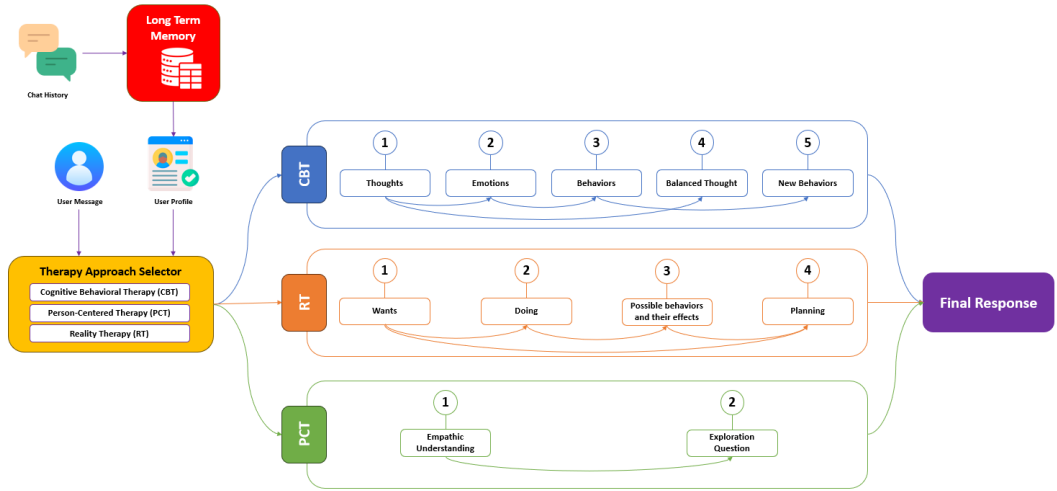

Figure 4: Overall architecture of the PsychoLexTherapy framework, integrating therapy approach selection, reasoning paths for CBT/RT/PCT, and long-term memory management for coherent multi-turn interaction.

Figure 4 illustrates an overview of the framework, showing how therapy reasoning paths were redesigned and integrated into the unified architecture. The following subsections detail the individual reasoning paths and their integration into the overall system.

3.4.1 Therapy Approach Selector Module

The Therapy Approach Selector functions as a decision-making layer that operates prior to the execution of therapeutic reasoning paths. Its role is to analyze the content of the user's message and select the most appropriate psychotherapeutic approach among the three integrated in PsychoLexTherapy: Cognitive Behavioral Therapy, Reality Therapy, and Person-Centered Therapy.

- When the message highlights automatic negative thoughts, cognitive distortions, or explicit signs of hopelessness, the system prioritizes a CBT reasoning path, which focuses on identifying and restructuring maladaptive cognitions.
- When the user primarily seeks unconditional acceptance, empathic listening, and emotional reflection, the module selects the PCT reasoning path, emphasizing empathy and client-led exploration.
- When the problem centers on unmet psychological needs, responsibility-taking, or counterproductive choices blocking goal attainment, the RT reasoning path is chosen, focusing on evaluating choices and aligning behavior with core needs.

By operationalizing these conceptual distinctions, the Therapy Approach Selector elevates the decision process from surface-level symptom matching to therapeutic reasoning, ensuring a more precise and contextually appropriate alignment between user input and intervention strategies.

*3.4.2 Cognitive Behavioral Therapy Reasoning Path*

Within PsychoLexTherapy, the Cognitive Behavioral Therapy reasoning path was explicitly designed to simulate therapeutic reasoning rather than surface-level empathetic response generation. The rationale is that for small language models to provide meaningful therapeutic support, they must follow a structured reasoning path that mirrors how a human therapist identifies, analyzes, and restructures maladaptive thought patterns.

The CBT reasoning path proceeds through six steps:

- Extraction of Automatic Thoughts – The SLM identifies short, distorted cognitions (e.g., "I always fail") that form the foundation of negative reasoning patterns.
- Inference of Emotional Consequences – From these cognitions, the system infers associated emotions (e.g., hopelessness, shame), simulating how a therapist reconstructs the client's inner state.
- Projection of Behavioral Tendencies – The model reasons about likely behavioral outcomes (e.g., withdrawal, irritability), linking cognition, affect, and behavior in a triadic structure central to CBT.
- Generation of Balanced Alternatives – Reasoning shifts toward counterfactual thinking, where the model formulates balanced, culturally sensitive alternatives that reduce distress.
- Derivation of Adaptive Behaviors – Based on revised cognitions, the framework proposes practical, culturally grounded behavioral strategies.
- Synthesis into a Therapeutic Response – The reasoning path is integrated into a natural, empathetic message that aligns with both linguistic fluency and therapeutic logic, while avoiding explicit clinical jargon.

By operationalizing these steps, PsychoLexTherapy transforms raw SLM outputs into structured reasoning trajectories. Instead of producing isolated empathetic sentences, the framework enables compact models to approximate the cognitive reasoning loop of CBT—a process essential for therapeutic coherence. This reasoning path illustrates how reasoning simulation bridges the gap between generic text generation and psychotherapy-aligned dialogue in resource-constrained settings.

*3.4.3 Reality Therapy Reasoning Path*

Within PsychoLexTherapy, the Reality Therapy (RT) reasoning path was developed to emphasize personal responsibility, conscious choice-making, and the fulfillment of fundamental human needs. The design encodes the reasoning process of a therapist into a sequence of structured steps that small language SLMs can simulate, ensuring responses go beyond generic empathy toward actionable therapeutic logic.

The RT reasoning path consists of five stages:

- Identification of Core Needs and Wants – The client's message is analyzed to extract their fundamental needs and underlying desires, such as respect, love, belonging, freedom, security, or belief in personal capability. This stage grounds the reasoning process in the motivational drivers that shape behavior.
- Analysis of Current Behaviors – The system examines the client's present behaviors and choices, determining which actions facilitate progress toward these needs and which act as barriers. This provides a snapshot of the client's current situation and sources of stagnation.
- Evaluation of Behavioral Consequences – Each identified behavior is assessed for its positive or negative outcomes, with particular attention to how effectively it contributes to or obstructs need fulfillment. This reasoning step helps the model replicate the therapeutic process of linking choices to consequences, reinforcing responsibility.

- Planning of Alternative Behaviors – Based on the evaluation, the framework proposes new, practical and culturally appropriate behaviors that can more effectively meet the client's needs. These suggestions emphasize feasibility, incremental change, and alignment with daily life contexts, encouraging sustainable progress.
- Integration into the Final Response – All reasoning steps are synthesized into a coherent, empathetic response. The final message acknowledges the client's needs, reflects on their behaviors, and outlines a clear pathway toward more effective and responsible choices. The tone is compassionate, non-judgmental, and culturally sensitive, while also fostering accountability and motivation for positive change.

Through this reasoning path, PsychoLexTherapy enables SLMs to produce responses that are not only natural and fluent but also therapeutically coherent and analytically grounded. Instead of superficial reassurance, the system guides the client toward recognizing responsibility for their choices and adopting strategies aligned with fundamental psychological needs.

*3.4.4 Person-Centered Therapy Reasoning Path*

The Person-Centered Therapy reasoning path in PsychoLexTherapy was designed to replicate the client-focused reasoning process of human therapists, where the emphasis lies on the client's inner experience, unconditional acceptance, and the facilitation of self-awareness and personal growth. Unlike directive approaches, PCT reasoning paths prioritize empathy and reflective listening, enabling SLMs to simulate the therapist's role as a supportive facilitator rather than a problem-solver.

The reasoning path unfolds in three stages:
- Empathic Reflection and Emotional Understanding – The client's message is analyzed to identify and reflect back key emotions (e.g., loneliness, insecurity, discouragement, shame) and core needs (e.g., love, acceptance, emotional safety). This empathic reflection assures the client that they are seen and heard, creating a safe environment for authentic self-expression.
- Exploratory Questioning for Self-Awareness – Following reflection, the system generates open-ended, compassionate questions that gently encourage deeper exploration of the client's experience. These questions are phrased in a supportive, non-judgmental manner to foster trust and to promote self-awareness, helping clients gain insight into their feelings, values, and unmet needs.
- Integration into a Final Supportive Response – All elements are synthesized into a warm, empathetic, and non-directive message that reflects the client's emotions and communicates unconditional positive regard. The response typically ends with an open-ended question, encouraging further exploration and ongoing self-discovery.

Through this design, PsychoLexTherapy enables SLMs to generate responses that are not only linguistically natural but also rich in emotional depth. By simulating the reasoning loop of PCT, the framework provides clients with an authentic experience of empathy, acceptance, and emotional safety, while guiding them toward self-reflection and personal growth.

*3.4.5 Long-Term Memory Management Module*

To maintain coherence and increase effectiveness in multi-turn interactions, the PsychoLexTherapy framework incorporates a Long-Term Memory Management Module. This module is inactive in single-turn settings, where messages are processed solely based on the current input. In contrast, for multi-turn sessions it is activated to provide continuity, consistency, and depth, enabling the system to sustain realistic therapeutic dialogue over extended conversations.

The memory module is implemented on top of the MemoBase system, which is responsible for creating and managing dynamic user profiles while enabling structured storage and retrieval of dialogue history. Instead of treating each message as an isolated unit, the framework can refer back to context and prior exchanges, generating responses that are more coherent, empathetic, and aligned with the user's previous disclosures.

The module performs several key functions:
- Automatic Profile Construction – MemoBase builds enriched, structured profiles from user interactions, spanning demographic details to personal preferences.
- Profiles as Memory Units – Profiles are organized hierarchically by topic and subtopic, designed for both machine usability and human readability.
- Data Storage – Information is stored as flexible data packages, allowing additions, retrievals, and deletions as needed.
- Buffering System – New information is first held in a temporary buffer before being transferred into the long-term profile, either automatically or with manual control.

Over time, this process produces comprehensive and evolving user profiles, enabling personalized and context-aware therapy simulations. As a result, the system delivers responses that are not only empathetic but also directly aligned with the user's history and emotional trajectory.

The structure of each user profile in the memory module consists of four main components:
- Basic Information – name, age, gender, occupational background, place of residence, spoken languages, and time zone.
- Ongoing Preferences – preferred conversation style, important personal topics, learning goals, and data storage preferences.
- Personalization Settings – inclinations toward humor or informal tone, and preferences for short or long responses.
- Recent Events – significant life changes extracted automatically from interactions, stored with temporal labels.

Through these mechanisms, the Long-Term Memory Management Module enables PsychoLexTherapy to produce deeper, context-sensitive, and personalized responses, offering an experience closer to genuine human interaction.

## 4 EVALUATION

### 4.1 Assessing Psychological Knowledge in Language Models

*4.1.1 Candidate Language Models*

For the first stage of evaluation, we selected open-source language models under 10 billion parameters. This threshold was chosen for both technical and ethical reasons. From a technical perspective, compact small language models can be executed on a single personal computer without requiring access to expensive cloud infrastructure or high-performance GPUs, thereby enhancing reproducibility and accessibility for researchers. From an ethical perspective, on-device deployment prevents the transfer of sensitive psychological data to external servers, safeguarding privacy and cultural confidentiality—a crucial concern in mental health applications, particularly for Persian data.

By restricting model size, we also achieved a balance between accuracy and computational cost. While larger models may demonstrate stronger raw performance, their hardware demands conflict with our research goal of creating a lightweight, reproducible, and privacy-preserving framework. Based on these considerations, we included models from the Gemma-3[21], Qwen-3[22], LLaMA-3.2[29], and Mistral families[30], covering a parameter range from ~1B to ~8B. This diversity allowed us to systematically analyze how different compact models represent and reason about psychological knowledge in Persian.

*4.1.2 PsychoLexEval Dataset and Evaluation Scenarios*

To probe the psychological competence of these models, we used the PsychoLexEval dataset. This dataset was purpose-built for evaluating psychological knowledge and consists of 3,430 multiple-choice questions, each with four options and exactly one correct answer. Items were drawn from graduate psychology entrance exams, professional recruitment tests, authoritative online resources, and LLM-generated questions inspired by Persian psychology textbooks.

Unlike open-domain corpora, PsychoLexEval is designed to reflect academic test conditions, forcing models to choose among plausible distractors. This setup not only measures linguistic processing but also tests a model's ability to select the most accurate conceptually grounded response. The questions span a broad range of psychological domains, from fundamental theories to applied clinical topics, enabling a nuanced assessment of theoretical depth.

Evaluation scenarios were constructed such that models were required to output exactly one option for each question. This constrained format facilitated the computation of standard metrics such as accuracy and allowed direct cross-model comparison under identical conditions.

*4.1.3 Experimental Configurations*

To ensure fairness and comparability, all models were evaluated under a uniform configuration. Text generation parameters were fixed across experiments, following common practices in LLM evaluation. Table 2 summarizes the key settings.

Table 2: Text generation parameters for psychological knowledge evaluation

| Parameter | Value | Description |
| --- | --- | --- |
| Temperature | 0.01 | Minimizes randomness to prioritize accuracy and consistency |
| Max tokens | 16 | Restricts output length to concise answer candidates |
| Top-p | 0.9 | Nucleus sampling for controlled probability distribution |

## 4.2 Baselines for Single-Turn Evaluation

To rigorously evaluate the proposed PsychoLexTherapy framework, we established multiple single-turn baselines. These baselines capture increasing levels of reasoning complexity, ranging from naive prompt-based empathy to structured multi-agent debate. By comparing PsychoLexTherapy against these alternatives, we aimed to isolate the added value of therapy-specific reasoning paths and framework-level integration beyond simple prompting strategies.

*4.2.1 Simple Baseline*

The simplest baseline used a generic empathetic prompt without any therapeutic reasoning modules. In this setting, the SLM generates responses that: (i) acknowledge the user's emotions, (ii) reflect them back in a warm, nonjudgmental tone, (iii) provide a culturally sensitive hopeful perspective, and (iv) optionally suggest small and practical coping steps. Importantly, this baseline explicitly avoids technical jargon or references to psychotherapy methods. It provides a reference point to measure whether structured reasoning significantly improves therapeutic alignment compared to a naive empathetic response generator.

*4.2.2 Simple Baseline + Therapy Selector*

The second baseline extends the simple prompt design with the Therapy Approach Selector module. Before generating a response, the selector analyzes the user's message and implicitly routes it to one of three therapy orientations: CBT, RT, or PCT. Although the final response remains empathetic and culturally adapted, it is additionally guided by the internal logic of the selected therapeutic orientation. For example, messages dominated by negative automatic thoughts are mapped to a CBT-oriented prompt; emotionally focused messages are routed to PCT; and responsibility- or choice-centered messages are routed to RT. This baseline evaluates the benefit of conceptual guidance without introducing explicit multi-step reasoning chains.

*4.2.3 Empathy Chain Baseline*

Inspired by prior work on empathy-enhanced chain-of-thought prompting[31], this baseline structures the response generation into two phases: a reasoning trace and a final empathetic message. In the reasoning phase, the model explicitly identifies key emotions, reconstructs possible situational triggers, and infers underlying needs or maladaptive cognitions. Only after this structured reasoning does the model generate a final, fluent response.

For implementation, three therapy-oriented variants were designed:
- CBT-variant: detects negative cognitions, identifies distortions, and reformulates them into balanced alternatives.
- PCT-variant: emphasizes empathic reflection and ends with an open-ended question to encourage deeper self-exploration.
- RT-variant: identifies unmet needs, evaluates current behaviors, and proposes small, actionable alternatives.

By separating reasoning from the final message, the Empathy Chain reduces the risk of shallow or repetitive outputs and ensures that responses remain anchored in therapeutic logic while still being natural and empathetic in delivery.

*4.2.4 Empathic Agents Baseline*

The most advanced baseline adopts a multi-agent debate structure [32]. Instead of a single SLM generating a response, multiple reasoning agents—each aligned with a distinct therapy orientation (CBT, RT, PCT)—produce candidate responses. Each agent explains its reasoning, critiques the others, and defends its therapeutic stance. A final decision agent then synthesizes the strongest elements of these responses into a unified output that is empathetic, culturally grounded, and free of explicit technical references.

This design mimics real-world therapeutic pluralism, where multiple schools of thought provide complementary perspectives. By simulating debate among therapy-oriented agents, this baseline generates responses that are often richer, more diverse, and less repetitive than single-model outputs. It thus provides a strong comparator for PsychoLexTherapy, isolating the contribution of integrated structured reasoning and long-term memory beyond multi-agent prompting alone.

**4.3 Baselines for Multi-Turn Evaluation**

To benchmark PsychoLexTherapy in multi-turn therapeutic dialogue, we implemented several baselines that vary in their ability to retain conversational context and simulate reasoning. These baselines establish reference points for measuring how structured reasoning paths and long-term memory integration contribute to therapeutic coherence, empathy, and personalization.

*4.3.1 Simple Multi-Turn Prompt*

The first baseline provides the model with the entire raw conversation history at each turn. Unlike the single-turn mode, the model can refer back to earlier messages, enabling a degree of topical continuity and emotional consistency. However, no structured mechanism is applied for information management: the dialogue history is concatenated verbatim, which becomes inefficient and error-prone as conversations grow longer. This baseline allows us to measure the extent to which naive history concatenation improves coherence compared to context-free generation.

*4.3.2 Multi-Turn Prompt with Long-Term Memory*

The second baseline introduces the Long-Term Memory Module (MemoBase) in addition to raw dialogue history. Instead of passing the full transcript, user information is selectively summarized and organized into dynamic profiles, which capture key dimensions such as dominant emotions, core needs, behavioral patterns, and recent life events. These profiles are updated continuously, allowing the model to adapt its responses in a personalized and context-aware manner. Compared to simple concatenation, this approach enhances efficiency, reduces repetition, and aligns responses more closely with the evolving emotional state of the user.

*4.3.3 PsychoLexTherapy without Long-Term Memory*

This intermediate version of PsychoLexTherapy integrates the therapy approach selector and structured reasoning paths (CBT, RT, PCT), but disables the Long-Term Memory Module. User inputs are processed based on raw dialogue history, without structured profiling or memory consolidation. As a result, responses benefit from therapeutic logic and reasoning coherence but may still fail to capture long-range continuity or personalize deeply to the user's evolving profile. This baseline allows us to isolate the contribution of structured therapeutic reasoning from the contribution of long-term memory.

*4.3.4 PsychoLexTherapy with Long-Term Memory*

The full PsychoLexTherapy system combines reasoning-oriented therapeutic paths with the Long-Term Memory Module, producing the most realistic simulation of therapy-like dialogue. At each turn, the Therapy Approach Selector determines the most appropriate reasoning path based on both the user's current message and their dynamic profile. The selected reasoning path (e.g., CBT for dysfunctional thoughts, RT for responsibility and needs, PCT for empathy and self-exploration) is then executed, guided by structured memory that encodes the user's history, preferences, and emotional trajectory. This design ensures both short-term coherence (via reasoning paths) and long-term personalization (via memory management). It represents the upper bound of our evaluation, against which all other baselines are compared.

## 4.4 Evaluation Setup

The evaluation of PsychoLexTherapy was conducted in two complementary settings: single-turn interactions, which tested the framework's ability to handle isolated user queries, and multi-turn dialogues, which examined its capacity to sustain therapeutic reasoning and coherence across extended sessions. Both settings were designed to reflect culturally grounded and clinically relevant scenarios in Persian psychotherapy.

*4.4.1 PsychoLexQuery Dataset and Single-Turn Testing*

For single-turn evaluation, we used the PsychoLexQuery dataset, which comprises approximately 4,000 real user questions collected from Persian-language psychological forums and counseling platforms. These questions span diverse themes, including family and marital conflicts, academic stress, workplace difficulties, social anxiety, depression, and self-esteem challenges.

Each question was treated as an independent input, and the system generated a response without access to prior conversational context. The Therapy Approach Selector module determined the most appropriate reasoning path (CBT, RT, or PCT) based on the content of each query, after which the corresponding reasoning pipeline produced the final response. To ensure comparability, all outputs were evaluated without human post-editing and rated by human judges according to predefined criteria.

*4.4.2 PsychoLexDialogue Dataset and Multi-Turn Testing*

To evaluate multi-turn interactions, we used the PsychoLexDialogue dataset, specifically developed for this study to simulate realistic therapeutic conversations. Unlike single-turn queries, each entry in this dataset was organized as a multi-stage dialogue scenario, reflecting the dynamic progression of therapy sessions. Dialogues covered the stages of therapeutic engagement: establishing rapport, emotional reflection, guided self-exploration, fostering growth, and concluding with goal-setting or reflection.

The dataset allowed testing under two key conditions:
- Without structured memory, where models relied solely on raw concatenated histories.
- With the Long-Term Memory Module (MemoBase), where user interactions were summarized into structured, dynamic profiles capturing emotions, needs, behavioral tendencies, and recent events.

This design provided a controlled environment to assess the contribution of structured memory in enhancing coherence, personalization, and emotional continuity across multiple conversational turns.

*4.4.3 Evaluation Metrics*

To ensure a comprehensive assessment of system performance, we defined separate evaluation criteria for single-turn and multi-turn interactions. All responses were rated on a ten-point Likert scale (1 = very poor, 10 = excellent) by human annotators.

For single-turn interactions, metrics emphasized the immediacy, clarity, and therapeutic alignment of individual responses:
- Empathy: Ability to recognize and reflect the user's emotional state in a warm, supportive, and non-judgmental manner.
- Coherence and Structure: Presence of a logically organized response with clear beginning, body, and closure.
- Cultural Fit: Alignment with Persian cultural and linguistic norms, ensuring contextual appropriateness.
- Therapeutic Alignment: Consistency with the reasoning path of the selected approach (CBT, RT, or PCT) without explicit use of clinical terminology.
- Content Accuracy: Specificity and relevance of the response to the user's input, avoiding vague or generic statements.
- Adaptability: Ability to personalize the response according to the user's expressed situation and needs.
- Linguistic Fluency: Grammatical correctness, naturalness, and smooth sentence construction.
- Clarity: Ease of understanding and absence of unnecessary ambiguity.

- Human-likeness: Degree to which the response resembles the style and tone of a human therapist.

For multi-turn interactions, additional dimensions were introduced to capture long-range coherence, personalization, and affective continuity across sessions:

- Empathy: Continuous recognition and validation of the client's emotions throughout the dialogue.
- Therapeutic Consistency: Adherence to the logic of the selected reasoning path across multiple turns.
- Continuity: Maintenance of topic progression and reference to prior exchanges, avoiding contradictions or resets.
- Emotional Consistency: Sensitivity to shifts in the client's emotional trajectory over the course of the dialogue.
- Personalization: Effective use of user history and dynamic profiling to tailor responses to individual circumstances.
- Cultural Fit: Appropriateness of responses within Persian sociocultural norms across extended contexts.
- Completeness: Coverage of all major elements raised by the user within and across turns.
- Linguistic Fluency and Cohesion: Natural, grammatically correct, and stylistically consistent language, with smooth transitions between turns.
- Clarity: Transparency and comprehensibility of responses without overcomplication.
- Variety: Use of diverse phrasing and strategies to avoid repetitive or formulaic replies.
- Human-likeness: Overall resemblance to the dynamics of real therapist–client dialogues, including warmth and natural pacing.
- Overall Preference: Annotators' global judgment of which response best addresses the user's needs among compared systems.

*4.4.4 Human and LLM-as-a-Judge Evaluation*

To ensure comprehensive assessment, two complementary evaluation procedures were adopted: automatic evaluation with LLM-as-a-judge and human evaluation with psychology graduate students. For both single-turn and multi-turn experiments, GPT-5 were employed as evaluators. Following recent practices in LLM evaluation, each system output was scored by an independent LLM on a five-point Likert scale across predefined criteria. The evaluator model was prompted with anonymized outputs and explicit scoring instructions to minimize bias. Automatic scoring enabled large-scale, consistent evaluation and allowed fine-grained comparison across all systems.

In addition to automatic evaluation, a comparative human preference study was conducted in the single-turn setting. Three graduate students in psychology served as blinded annotators. For each sampled query, the raters were presented with five anonymized system outputs (from all baselines and PsychoLexTherapy) and were asked to assign a strict overall ranking (1 = best, 5 = worst) with no ties permitted. Unlike the LLM-as-a-judge protocol, annotators did not score each criterion separately; instead, they used the predefined criteria as guidelines for their comparative judgment.

This dual protocol provided both fine-grained automatic scores and human-grounded overall preference rankings. The combination allowed us to validate PsychoLexTherapy's performance from two perspectives: large-scale LLM-based evaluation across single-turn and multi-turn settings, and targeted human preference validation in the single-turn context.

# 5 RESULTS AND DISCUSSION

## 5.1 Psychological Knowledge Evaluation Results

To assess the extent of psychological knowledge encoded in small language models, we employed the PsychoLexEval dataset in a zero-shot setting. In this scenario, models were required to answer multiple-choice psychology questions without exposure to in-context examples or fine-tuning, relying solely on their internal knowledge representations. The primary metric of evaluation was accuracy, defined as the percentage of correctly answered questions.

As summarized in Table 3, the results reveal a substantial variance across model families. The Gemma-3 and Qwen-3 series consistently outperformed other candidates, achieving accuracies of 55.2% and 53.0% respectively at the 7–8B parameter scale, and maintaining moderate performance in their 4B variants (50.4% and 48.3%). In contrast, smaller models such as LLaMA-3.2 (1.2B, 3.2B) and Mistral-7B demonstrated markedly lower accuracies (21–31%), suggesting limited psychological knowledge and reduced suitability for downstream therapeutic reasoning.

Based on these findings, the Gemma-4B model was selected as the base model for subsequent stages of PsychoLexTherapy. This decision balanced three critical factors: (i) sufficient psychological competence for baseline reasoning, (ii) computational feasibility for execution on consumer-grade hardware, and (iii) privacy preservation through fully local deployment without reliance on external servers. These properties made Gemma-4B an optimal candidate for building a practical, reproducible, and privacy-conscious therapeutic framework in Persian.

Table 3: Accuracy of Candidate Models on the PsychoLexEval Dataset

| Model | Parameters | Accuracy (%) |
| --- | --- | --- |
| Gemma-3 | 7.8B | 55.2 |
| Qwen-3 | 8.2B | 53 |
| Gemma -3 | 4.3B | 50.4 |
| Qwen -3 | 4.0B | 48.3 |
| Gemma -3 | 1.0B | 33.1 |
| Mistral | 7.2B | 31.2 |
| LLaMA-3.2 | 3.2B | 28.7 |
| LLaMA-3.2 | 1.2B | 21.3 |

## 5.2 Automatic Evaluation Results in Single-Turn

To evaluate PsychoLexTherapy in single-turn settings, its outputs were compared against four baselines on the PsychoLexQuery dataset using LLM-as-a-judge scoring across psychological, therapeutic, and linguistic criteria. The results are reported in Table 4.

The overall pattern shows that the more structurally grounded and therapy-oriented the design, the higher the scores across empathy, coherence, cultural fit, and content accuracy. PsychoLexTherapy achieved the strongest performance, Empathic Agents ranked second, the Empathy Chain displayed intermittent but unstable improvements, and the two simpler baselines trailed behind. This trend aligns with the design logic of the systems: approaches that incorporate explicit therapeutic decision-making, stepwise reasoning, or multi-agent debate outperform generic prompt-based methods, which tend to drift toward clichés or shallow suggestions.

Table 4: Automatic single-turn evaluation results (LLM-as-a-Judge) on the PsychoLexQuery dataset

| Method | Empathy | Coherence & Structure | Cultural Fit | Therapeutic Alignment | Content Accuracy | Adaptability | Fluency | Clarity | Human-likeness | Mean |
|---|---|---|---|---|---|---|---|---|---|---|
| Simple Prompt | 3.12 | 2.92 | 4.42 | 1.24 | 1.9 | 0.42 | 3.26 | 5.21 | 2.50 | 3.15 |
| Simple + Therapy Selector | 3.96 | 4.16 | 5.62 | 2.5 | 0.42 | 2.08 | 5.62 | 5.42 | 3.33 | 3.67 |
| Empathy Chain | 2.43 | 1.62 | 2.92 | 1.7 | 1.44 | 1.44 | 2.71 | 1.88 | 3.95 | 6.19 |
| Empathic Agents | 5.42 | 7.08 | 6.88 | 7.92 | 6.24 | 6.24 | 6.67 | 6.46 | 5.21 | 6.45 |
| **PsychoLexTherapy** | **6.25** | **8.76** | **7.29** | **8.34** | **7.08** | **7.5** | **7.08** | **6.88** | **6.04** | **7.24** |

In the Simple Prompt baseline, the main strength lies in the fluent and warm language style. Because the model directly generates responses without any intermediate reasoning, its outputs are grammatically smooth and easy to read. However, the absence of mechanisms for systematically identifying therapeutic needs, reconstructing thoughts or emotions, or planning interventions leads to weak therapeutic alignment and content accuracy. Responses are often generic and under-personalized, especially in complex cases, where lack of structure also reduces coherence and clarity.

The Simple plus Selector baseline partly mitigates these weaknesses: once the therapy selector infers the most suitable approach, the tone and content become more aligned with therapeutic logic, improving accuracy and alignment. However, its reasoning depth remains shallow, with responses often reduced to surface-level heuristics. While the text appears more therapy-oriented in form, it remains vague and sometimes inconsistent with the user's lived experience.

The Empathy Chain baseline, in theory, should resolve many of these shortcomings by forcing the model to perform stepwise reasoning—detecting emotions, reconstructing context, identifying needs or thoughts, and then generating a response. When executed properly, this chain improves content accuracy and inferential clarity, yielding deeper responses. Yet its dependence on chain execution introduces volatility: if the reasoning steps are long but poorly compressed, final responses become verbose, repetitive, or mechanical, lowering fluency and coherence. Conversely, when constrained by short output length, the reasoning chain may be truncated, leading to incomplete or ambiguous responses. This explains the unusual scores in the "human-likeness" dimension: in some outputs, the analytical tone intrudes, making the responses appear unnatural to raters.

The Empathic Agents baseline demonstrates substantial improvement. Diversity of perspectives and mutual critique among agents allow blind spots to be covered, complementary ideas to be combined, and the decision-maker to synthesize the strongest segments. This results in empathy being conveyed from multiple angles (emotional reflection, cognitive reframing, small behavioral suggestions), while simultaneously improving cultural fit and content accuracy. Its main weakness is potential verbosity and polyphony: without robust aggregation, final outputs may become long, repetitive, or occasionally inconsistent in tone or recommendations.

Finally, PsychoLexTherapy combines three distinct advantages. First, predefined therapeutic reasoning paths organize key decisions before text generation, ensuring systematic alignment with CBT, RT, or PCT principles. Second, cultural and stylistic constraints for Persian reduce risks of mismatched tone, inappropriate forms of address, or culturally incongruent examples. Third, built-in quality control checklists prevent the drift toward clichés or scattered recommendations. Together, these mechanisms raise therapeutic alignment, accuracy, and clarity while preserving human-likeness and fluency by hiding intermediate reasoning from the surface output. Potential risks arise when inputs fall outside the distribution of the designed rules; in such cases, protective mechanisms may yield overly cautious responses.

## 5.3 Human Evaluation Results in Single-Turn

In addition to automatic LLM-as-a-judge evaluation, we conducted a human evaluation on the PsychoLexQuery dataset to further assess the performance of PsychoLexTherapy against the four baselines. Three graduate students in psychology served as annotators. For each query, they were asked to rank the system outputs comparatively rather than score them on individual criteria, ensuring a clear preference ordering. The results are summarized in Table 5.

The human evaluation results are fully consistent with the automatic judgments: PsychoLexTherapy was rated highest overall, while the Simple Prompt baseline performed worst. This alignment strengthens confidence that the proposed framework's advantages are both quantitatively measurable and qualitatively recognizable to human experts.

Table 5: Human evaluation (single-turn) of PsychoLexTherapy and baselines (mean rank, lower is better)

| Method | Mean Rank |
|---|---|
| Simple Prompt | 3.25 |
| Simple Prompt + Therapy Selector | 3.20 |
| Empathy Chain | 3.16 |
| Empathic Agents | 2.00 |
| **PsychoLexTherapy** | **1.43** |

Among the baselines, both the Empathy Chain and the Empathic Agents approaches provided substantial improvements over the simpler prompts. The Empathy Chain encouraged the model to engage in stepwise analytical reasoning before generating responses, which often improved content accuracy and inferential clarity. However, its heavy reliance on proper chain execution sometimes led to unstable outputs: overly analytical or mechanical tones when reasoning was verbose, or incomplete answers when reasoning was truncated.

By contrast, the Empathic Agents baseline produced richer and more diverse responses by combining multiple therapeutic voices. Each agent compensated for the weaknesses of others, and the decision-maker synthesized a more comprehensive final response. The trade-off was occasional verbosity or redundancy due to polyphony, but overall, this approach outperformed the simpler baselines.

Interestingly, the Simple + Therapy Selector baseline was judged by humans as nearly indistinguishable from the plain Simple Prompt. Two reasons explain this: (i) if the selector misclassified the therapeutic approach or interpreted cues superficially, the output seemed misaligned to raters; and (ii) even when the correct approach was chosen, without deeper reasoning structure the output remained advice-like and generic, offering little perceptible improvement over a plain empathetic prompt. This suggests that merely labeling responses with therapeutic orientation is insufficient—structured reasoning paths are essential.

Human raters consistently favored PsychoLexTherapy for two main reasons: (1) its stepwise therapeutic workflows, which ensured that responses were purposeful and directly relevant to the user's issue; and (2) its linguistic and cultural grounding, which preserved a natural, fluent, and stereotype-free tone. These qualities were precisely what raters associated with overall response quality, explaining why PsychoLexTherapy achieved the top rank across nearly all test cases.

## 5.4 Automatic Evaluation Results in Multi-Turn

To evaluate PsychoLexTherapy in multi-turn settings, we compared its outputs on the PsychoLexDialogue dataset against four baselines:

- Multi-turn without memory (access to raw conversation history only)
- Multi-turn with long-term memory (structured profile-based summarization)
- PsychoLexTherapy without long-term memory
- PsychoLexTherapy with long-term memory (full version)

Responses were scored by an LLM judge using the defined 10-point scales. Results are summarized in Table 5, which reports scores across both therapeutic and linguistic dimensions.

Table 5: Comparison of Baselines and PsychoLexTherapy Variants on in Multi-Turn Evaluation

| Method | Empathy | Cultural Fit | Therapeutic Alignment | Content Accuracy | Adaptability | Content Coherence | Emotional Coherence | Personalization | Fluency | Linguistic Coherence | Clarity | Diversity | Style Consistency | Linguistic Human- | Average |
|---|---|---|---|---|---|---|---|---|---|---|---|---|---|---|---|
| Multi-Turn w/o Memory | 7.8 | 8 | 6.8 | 7.2 | 7 | 6.8 | 6.6 | 6.4 | 3.1 | 3.3 | 3.5 | 2.9 | 3.2 | 3.4 | 5.43 |
| Multi-Turn + Memory | 8.4 | 8.4 | 7.8 | 7.8 | 7.6 | 7.8 | 7.6 | 7.4 | 4.2 | 4.5 | 4.3 | 4 | 4.4 | 4.6 | 6.34 |
| PsychoLexTherapy w/o Memory | 8.6 | 8.6 | 8.2 | 8.2 | 8 | 7.8 | 7.8 | 7.6 | 5.6 | 5.3 | 5.8 | 5.1 | 5.5 | 5.7 | 6.99 |
| **PsychoLexTherapy + Memory** | **9.2** | **9** | **8.8** | **8.6** | **8.4** | **8.8** | **8.6** | **8.6** | **7.3** | **7.5** | **7.1** | **7** | **7.4** | **7.6** | **8.14** |

The results reveal a clear progression: as models move from simple multi-turn prompting toward the full PsychoLexTherapy framework, performance improves across nearly all criteria.

- Multi-turn without memory: Having access to raw history improves over single-turn prompting, but the absence of structured memory often leads to incoherence, omissions, or attention to irrelevant details as dialogues grow longer.
- Multi-turn with long-term memory: Structured summarization significantly improves personalization and content continuity, as responses can reference prior user states and maintain emotional coherence.
- PsychoLexTherapy without memory: Even without memory, the structured therapeutic reasoning paths enable stronger therapeutic alignment and content accuracy, producing more logically grounded and culturally appropriate responses. However, this version cannot track long-term changes in user states.
- PsychoLexTherapy with memory (full version): Combining structured therapeutic workflows with long-term memory yields the best results. The framework achieves high ratings in empathy, personalization, fluency, and clarity, reflecting interactions that closely resemble real therapeutic sessions.

In summary, the findings show that long-term memory enhances personalization and coherence, while therapeutic reasoning paths improve accuracy and alignment. Their integration in the full framework produces the most consistent and human-like experience, surpassing all baselines in multi-turn evaluation.

## 6 LIMITATIONS AND FUTURE WORKS

This study faces several limitations that should be acknowledged when interpreting the findings. First, we intentionally constrained all candidate models to fewer than 10 billion parameters to ensure on-device feasibility and protect user privacy. While this decision aligned with the research goals, it inevitably reduced absolute accuracy on knowledge-intensive items and may underrepresent the full therapeutic reasoning capacity of larger-scale models. Second, the knowledge evaluation was conducted in a strict zero-shot regime using the PsychoLexEval dataset. Alternative setups such as few-shot prompting, retrieval-augmented generation, or tool-assisted reasoning were not explored, and these might yield different rankings and insights into model capabilities.

A further limitation concerns the scope and cultural specificity of the datasets. All three PsychoLex corpora were developed in Persian and deliberately grounded in Iranian cultural contexts. This strengthens internal validity but narrows external generalizability; extending the framework to other languages or cultural settings requires careful adaptation and validation. Additionally, parts of the PsychoLexDialogue dataset were constructed through hybrid methods that included LLM-assisted components. Despite multiple layers of human review, such data may still carry subtle stylistic artifacts or synthetic biases that could influence multi-turn outcomes.

From a therapeutic standpoint, the reasoning paths were limited to three classical approaches—Cognitive Behavioral Therapy, Reality Therapy, and Person-Centered Therapy. Other modalities, such as Acceptance and Commitment Therapy (ACT), Emotion-Focused Therapy (EFT), schema therapy, or mindfulness-based approaches, were not modeled, potentially limiting the range of therapeutic strategies reflected in system responses. Similarly, although the MemoBase module enabled structured and persistent memory, it did not implement advanced mechanisms for forgetting, uncertainty handling, or conflict resolution, which may lead to profile drift over long-term interactions.

The evaluation methodology also introduces constraints. Human judgments, while performed by trained and blinded raters, remain subject to variability and construct ambiguity. Moreover, no clinical outcome measures (e.g., PHQ-9 or GAD-7 symptom change) or therapeutic alliance scales (e.g., WAI) were employed in a prospective setting, meaning the framework's impact on real-world therapeutic outcomes remains untested. Safety and clinical boundaries constitute another major limitation: the system is not a certified medical device, lacks validated crisis-detection or escalation protocols, and has not undergone randomized controlled trials (RCTs).

Finally, the results demonstrate sensitivity to prompt design and configuration choices. Variations in template wording, decoding parameters, or model checkpoints could lead to different performance profiles, and our ablation experiments were not exhaustive. Collectively, these limitations highlight the need for cautious interpretation of the findings and underscore opportunities for further refinement and clinical validation in future research.

# 7 CONCLUSION

This study introduced PsychoLexTherapy, a lightweight and culturally grounded framework designed to simulate psychotherapeutic reasoning with small language models. By systematically combining structured reasoning paths for Cognitive Behavioral Therapy, Reality Therapy, and Person-Centered Therapy with a long-term memory module (MemoBase), the framework advances beyond surface-level empathetic text generation and enables interpretable, therapy-aligned dialogue.

To support evaluation, we constructed the PsychoLexEval, PsychoLexQuery, and PsychoLexDialogue datasets, covering psychological knowledge assessment, single-turn response generation, and multi-turn conversational reasoning. Together, these resources provided a reproducible pipeline for benchmarking therapeutic dialogue systems in Persian—a language and cultural context that remains underrepresented in digital mental health research.

Empirical results showed that PsychoLexTherapy consistently outperformed baseline methods across psychological, linguistic, and cultural metrics. In particular, the integration of structured reasoning and long-term memory significantly improved coherence, empathy, and personalization in multi-turn interactions. These findings highlight the potential of compact, privacy-preserving SLMs to deliver clinically meaningful support in resource-constrained and culturally specific environments.

At the same time, we acknowledge that PsychoLexTherapy is a research prototype rather than a deployable clinical tool. Its current scope is limited to three therapeutic modalities, its datasets remain Persian-centric, and rigorous clinical outcome studies (e.g., randomized controlled trials) are necessary before real-world adoption.

In sum, this work contributes a novel framework, culturally aligned datasets, and empirical insights into the feasibility of simulating therapy-oriented reasoning with small models. We hope that it provides a foundation for future research at the intersection of language models, psychotherapy, and culturally sensitive digital health.

## ETHICS STATEMENT

This study was conducted in accordance with institutional and national ethical guidelines. The dataset (PsychoLexEval, PsychoLexQuery, and PsychoLexDialogue) was constructed from publicly available Persian mental health forums and anonymized prior to analysis to protect user privacy. No personally identifiable information (PII) was collected or stored. Human evaluation was carried out with three psychology graduate students, who provided informed consent prior to participation. According to the policies of Iran University of Science and Technology, this study was deemed exempt from full ethics board review because it involved only secondary analysis of publicly available, anonymized data and minimal-risk human evaluation.

**Declaration of generative AI and AI-assisted technologies in the manuscript preparation process**

During the preparation of this work the author(s) used OpenAI ChatGPT in order to assist with language polishing, text summarization, and structural refinement. After using this tool, the author(s) reviewed and edited the content as needed and take full responsibility for the content of the published article.